\documentclass[journal]{IEEEtran}
\IEEEoverridecommandlockouts    

\usepackage{graphicx} 
\usepackage[hidelinks]{hyperref}
\usepackage[dvipsnames]{xcolor}
\usepackage{tikz}
\usepackage[export]{adjustbox}
\usepackage{subcaption}
\usepackage{algorithm2e}
\usepackage{makecell}
\usepackage{array}
\usepackage[margin=1in]{geometry}
\usepackage{amsmath}

\title{\LARGE \bf

RoboTwin: A Robotic Teleoperation Framework Using Digital Twins
}

\author{
Harsha Yelchuri$^{1*}$,
Diwakar Kumar Singh$^{2*}$,
Nithish Krishnabharathi Gnani$^{3*}$,
T V Prabhakar$^{4*}$,
Chandramani Singh$^{5*}$
\thanks{%
*Department of Electronic Systems Engineering, Indian Institute of Science, Bengaluru, India\\
\textsuperscript{1}\texttt{harshayelchuri2000@gmail.com}\\
\textsuperscript{2}\texttt{singhdiwakar9713@gmail.com}\\
\textsuperscript{3}\texttt{nithishkgnani@gmail.com}\\
\textsuperscript{4}\texttt{tvprabs@iisc.ac.in}\\
\textsuperscript{5}\texttt{chandra@iisc.ac.in}%
}
}

\begin{document}

\maketitle
\thispagestyle{empty}
\pagestyle{empty}

\begin{abstract}
Robotic surgery imposes a significant cognitive burden on the surgeon. This cognitive burden increases in the case of remote robotic surgeries due to latency between entities and thus might affect the quality of surgery. Here, the patient side and the surgeon side are geographically separated by hundreds to thousands of kilometres. Real-time teleoperation of robots requires strict latency bounds for control and feedback. We propose a dual digital twin (DT) framework and explain the simulation environment and teleoperation framework. Here, the doctor visually controls the locally available DT of the patient side and thus experiences minimum latency. The second digital twin serves two purposes.  Firstly, it provides a layer of safety for operator-related mishaps, and secondly, it conveys the coordinates of known and unknown objects back to the operator's side digital twin. We show that teleoperation accuracy and user experience are enhanced with our approach. Experimental results using the NASA-TLX metric show that the quality of surgery is vastly improved with DT, perhaps due to reduced cognitive burden. The network data rate for identifying objects at the operator side is 25x lower than normal. 

\end{abstract}


\begin{figure*}[t]
    \centering
    \begin{subfigure}[t]{0.53\textwidth}
        \centering
        \includegraphics[width=\textwidth]{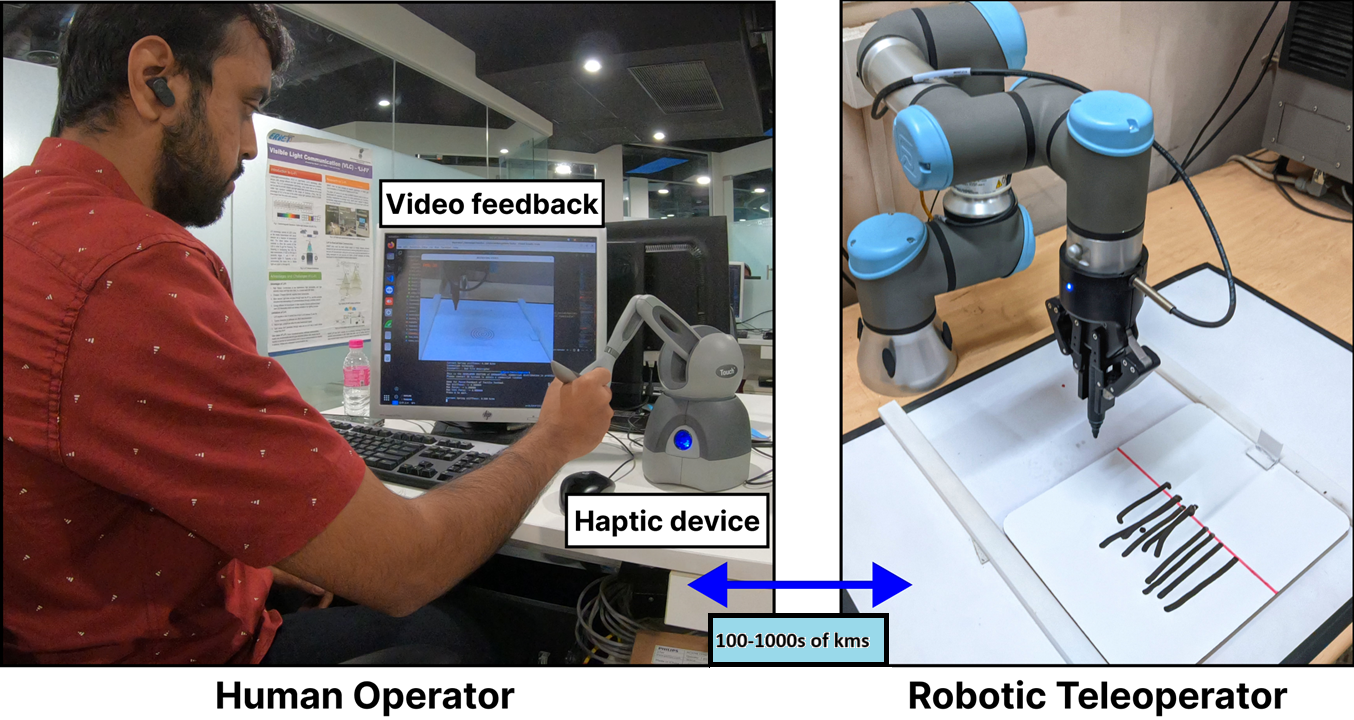}
        \caption{Testbed }
        \label{fig:test1a}
    \end{subfigure}%
    \hfill
    \begin{subfigure}[t]{0.38\textwidth}
        \centering
        \includegraphics[width=\textwidth]{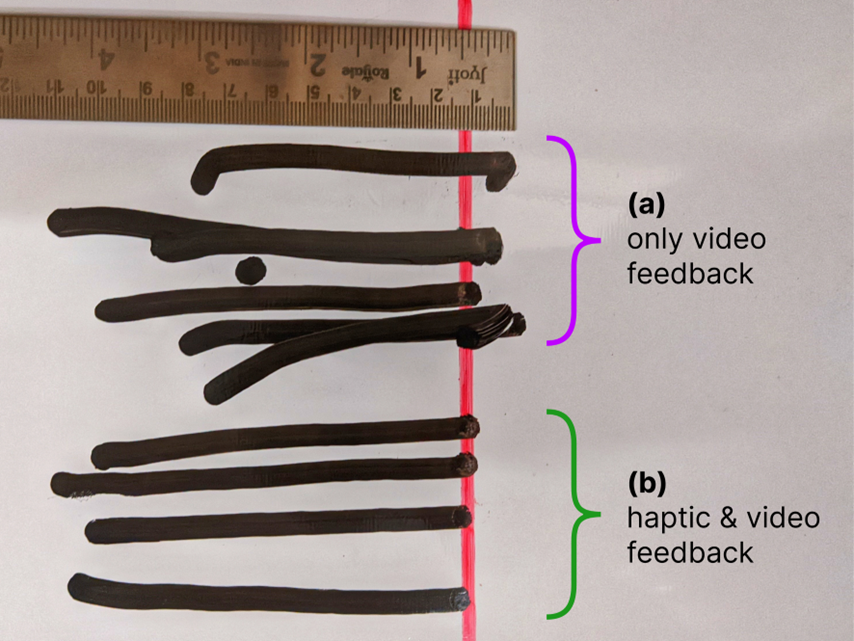 }
        \caption{Advantage of haptic feedback}
        \label{fig:test1b}
    \end{subfigure}
    \caption{Existing testbed in robotic teleoperation}
    \label{fig:testbed}
\end{figure*}
\section{Introduction and Motivation}

Minimally invasive robotic surgery systems, such as the SSi Mantra \cite{ssimantra2024} and the DaVinci system \cite{freschi2013technical}  offer a highly magnified and 3D view of the area with manipulation to perform precise surgery. The key components of a robotic surgery are (a) Surgical robot (like da Vinci) located at the patient side.  This system carries out all the physical tasks: cleaning, cutting, suturing, etc. This system is usually a slave of the surgeon. (b) The surgeon console is an interface that the surgeon uses to manipulate the robotic system. There is a screen that live streams the patient side infrastructure with a 3D view. (c) The network infrastructure that connects the patient side with the surgeon console. (d) In addition to the surgical robot, there are several sensors and associated interfaces located on both sides. For instance, there can be cameras, haptic devices, force torque sensors, contact sensors, etc. In robotic surgery, the surgeon must control complex robotic systems while simultaneously interpreting visual feedback, monitoring instrument response, and maintaining awareness of the surgical environment. This multitasking can impose a significant cognitive burden, leading to increased mental workload, fatigue, and a higher risk of errors.

While robotic surgery systems are well-established, the surgical robot and surgical console are co-located. This constraint and the limited availability of skilled surgeons make remote robotic surgery a viable solution.  Remote robotic surgery, also known as telesurgery$/$cybersurgery, allows the surgeon to use the surgical console from a significantly large distance. It is placed remotely over hundreds or thousands of $Km$s away from the surgical robot. This arrangement would alleviate the need for expert surgeons to travel between hospitals located across states or even continents. For safety, the surgeon console display is duplicated at the remote side. The network infrastructure is the backbone that carries bidirectional data. This includes on-premises network equipment such as switches and routers that usually connect to the service provider's network. A 5G network is envisaged for minimal latency.

For the success of remote robotic surgery, several challenges need to be addressed. Geofencing, 3D view, micro and macro movements, and tremor suppression are some of the requirements to ensure a safe and high-quality surgery. One important aspect is the surgeon's comfort level in terms of action, followed by instrument feedback, which should have a fast response. Even a slight discomfort may lead to increased cognitive load, resulting in cyber-sickness. Factors such as latency in control, limited haptic feedback, and non-intuitive interfaces further contribute to the surgeon’s cognitive load.  One such example is when the surgeon relies only on video feedback. Studies in literature \cite{gnani2024edgep4}, \cite{pal2024testbed} have conducted experiments to show that an operator's comfort level and accuracy improve with haptic feedback.
Fig. \ref{fig:testbed} shows the remote robotic surgery testbed. As earlier mentioned, it comprises (a) a human operator or surgeon, (b) a surgical console that houses the haptic device and the display of the remote site, and (c) the surgical robot with a depth camera that live streams the scene to the operator side.
The two sites are connected over a long distance. Fig. \ref{fig:test1a} highlights the current state of the art in remote robotic surgeries. Relying only on video feedback has the problem of inaccuracy in the surgery. Fig. \ref{fig:test1b}(a) shows that the vertical threshold line is exceeded by about 1 cm when the distance between the operator and teleoperator is about 400 km. Clearly this inaccuracy will increase with an increase in distance. The reason for this inaccuracy may be attributed to the delay in video transmission and image reconstruction at the operator side. This experiment was conducted with sufficient end-to-end bandwidth and therefore assumes abundant availability of network resources. Whereas in Fig. \ref{fig:test1b}(b) with haptic and video feedback, the threshold is not exceeded. The authors developed an operator side edge module to provide haptic feedback to the operator and arrest the hand movement exactly at the vertical threshold. Thus, deploying edge intelligence in the network significantly reduces the cognitive burden on the human operator.
 
 The Tactile Internet (TI) offers the necessary low latency and high reliability for effective teleoperation. Using TI, input actions from the operator reach the remote side, and feedback from the remote side promptly returns to the operator. A time-sensitive or low latency network with zero packet queuing delay should be the ideal platform to perform these surgeries. A teleoperation testbed deployed between two cities~\cite{pal2024testbed} with Time Sensitive Networking and Deterministic Networking implementation measured a round trip time (RTT) of about $10~ms$ and a video feedback latency of around $40~ms$ over a distance of $400~km$.

The current 5G and other upcoming standards specify service requirements for such cutting edge applications. For example, 3GPP TS 22.104 V19.2.0 (2024-06) specifies service requirements for cyber-physical control applications~\cite{3gppts}. Accordingly, remote surgery applications should achieve an end-to-end latency below $20~ms$ (practically a distance of $1000~km$) for a user equipment density of less than two devices per $1000~km^{2}$.

To overcome these limitations, operators can interact with a local digital twin of the remote teleoperator and its environment. This approach ensures that the user experience remains consistent regardless of the distance between the operator and the teleoperator. Digital twins facilitate the development and testing of robotic systems in a virtual world, reducing the time and cost associated with physical prototyping. Furthermore, using a digital twin enhances safety, allowing operators to address unexpected events before the real robot encounters them.

Our contributions are as follows: 
\begin{enumerate}
\item A  digital twin framework for drastic reduction in human cognitive load is proposed and verified.
\item A dual digital twin for significant reduction in network traffic between the surgeon (a.k.a. operator side) and patient side (a.k.a. remote side) is implemented.
\item A motion scaling algorithm to improve the quality of surgical procedures is demonstrated.
\item  A novel mechanism to reliably detect and display known and foreign objects at remote side is shown.

\end{enumerate}
\begin{figure*}[t]
    \centering
    \includegraphics[width=\textwidth]{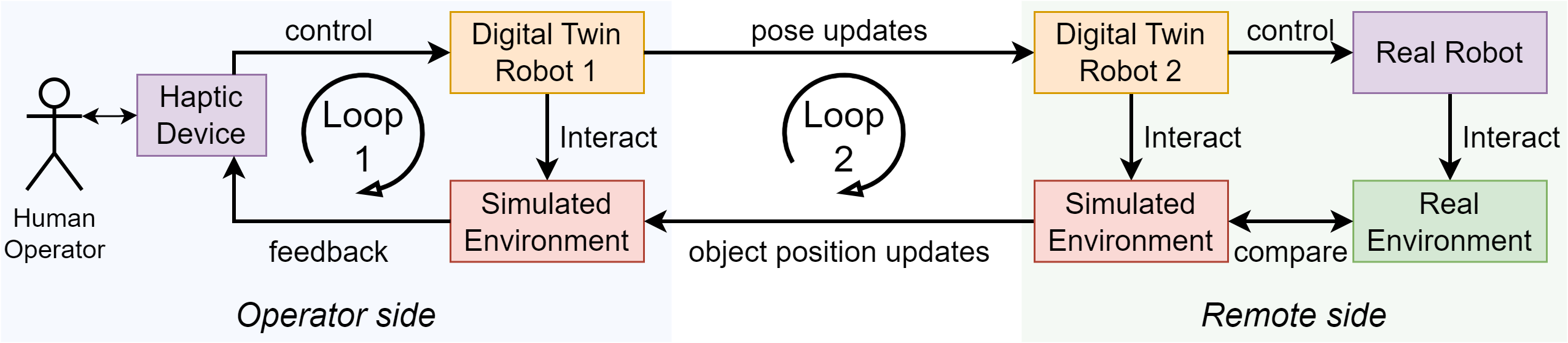}
    \caption{The twin loop architecture}
    \label{loops1}
\end{figure*}

\section{Related Work}
The application of medical digital twins hitherto is limited to real-time monitoring and control of a few organs, such as an artificial pancreas and a cardiac digital twin, with a goal to provide personalized medicine, such as precision medicine and intervention. However, in our case, humans enter the loop and interact in real time with process digital twins with various components, assets, and units linked and working together between the digital environment and the physical infrastructure. While process twins are widely applied in industrial settings, there is scanty literature available in the medical domain. Thus, their application in the medical domain is observed as a wide gap, particularly when remote operations are envisaged.

Several works in literature \cite{alkhayer2020}, \cite{jcm12165203}, \cite{LONGEAC2021386} show that surgeons used digital twins to simulate complex procedures in a virtual environment, allowing them to visualise the anatomy and identify potential challenges and thus refine their approach. The majority of the tools used include AR/VR glasses and other medical simulation software. For example, in jaw correction surgeries, software tools such as SimPlant \cite{simplant2025} and Dolphin \cite{dolphin2025} are popular.
There is no mention of the surgical procedure or process using twins as well as haptic feedback to surgeons. The authors in \cite{jcm14020324} describe in detail the application of DT in oncology, cardiology, neurology, and other healthcare areas. The authors cite several examples where the observational research and randomized clinical trials all show better results when DTs are used. Further, they mention that DTs have potential in lowering surgical complications and enhancing patient outcomes. 
In \cite{HALEEM202328} among several listed digital twin applications, there is a mention of a digital twin of a hospital. Stakeholders such as doctors, nurses, managers, etc. can obtain real-time insights regarding patients and processes.
In \cite{jpm14111101} the author  highlights the potential of Digital Human Twins towards personalized treatment such as precision medicine and intervention.
In \cite{jpm13101522} the challenges in creating the digital twin of an individual patient are discussed. Specifically, artificial pancreas and cardiac digital twin are discussed in detail.

Concerning testbeds, in \cite{Bzostek1999Testbed} a testbed for a robotically assisted and image-guided system for percutaneous delivery of surgical devices and therapeutic agents in the treatment of liver cancer and other malignancies is discussed. The authors rely on preoperative 3D images such as CT and MRI for planning the treatment pattern. The testbed does not support feedback, and latencies are zero due to the collocation of operator side and robot. 
The remote robotic surgery digital twin process testbed discussed in this paper has features to test robotic procedures and techniques in a controlled environment with a goal to evaluate robotic performance, operator satisfaction, and other control algorithms. 

\section{System Architecture}

To facilitate the intervention of a human operator in teleoperation and manipulate the real robot, Fig. \ref{loops1} shows our proposed architecture, which comprises a pair of digital twins, Digital Twin Robot 1 (DT Robot 1) and Digital Twin Robot 2 (DT Robot 2). These robots are available for manipulation and interaction in the simulated environment. The DT Robot 1 resides at the operator side and is part of a control and feedback loop called loop 1.  It has the advantage of following the human operator with almost negligible latency. For example, when the operator side  haptic device performs a gesture, the same is accomplished by the DT Robot 1 in real time. The DT Robot 1 sends operator action updates to DT Robot 2 and in turn receives updates from the remote side. DT Robot 2 is in real time coordination and collocated with the real robot which is part of loop 2.  Needless to say, loop 1 $\leftrightarrow$ loop 2 have a latency commensurate to the network latency.
A summary of the two loops is as follows:
\textbf{Loop 1:} operator $\leftrightarrow$ DT Robot 1 allows the operator to interact with DT Robot 1 and receive immediate haptic feedback. \textbf{Loop 2:} DT Robot 1 $\leftrightarrow$ remote side, ``pose'' comprising of the robot end effector coordinates and open/close status (x, y, z, o, c) from the DT Robot 1 is received by the DT Robot 2, and after applying inverse kinematics, it transmits joint position commands to the real robot and updates DT Robot 1 and DT Robot 2 with real-time changes in the remote environment to take care of discrepancies. DT Robot 2 plays a crucial role in facilitating this communication and synchronization.
Using this basic two-loop architecture, we evaluate several teleoperation features such as safety, motion scaling, pan and zoom, and known and foreign object detection. 
\subsection{Testbed Hardware}

The system operates through two primary feedback loops. The hardware for loop 1 comprises a Linux host PC with an AMD Ryzen 9 5900x 12-core processor and about 60 GB of usable RAM running Ubuntu 22.04. The haptic device used by the human operator is sourced from 3D Systems Geomagic Touch \cite{3dsystems_touch} with 6 degrees of freedom. It is interfaced using USB 2.0 to the PC. The OpenHaptics API is used to program the haptic device to perform operations and provide \text{(x, y, z)} coordinates as well as \text{(open, close)} functions. This PC also houses an Nvidia GeForce RTX 3090 GPU, and the digital twin is constructed using Nvidia's Isaac Sim \cite{nvidia_isaac_sim} software. 

The hardware for loop 2 comprises a Linux host PC with an Intel Core i9-10900K CPU and about 60 GB of usable RAM running Ubuntu 22.04. This PC has two GPUs, Nvidia GeForce RTX 3080 and 3090 GPUs. An RGB-depth camera capturing the real environment is interfaced to the PC using USB 3.2. The real robot, Universal Robot 3 (UR3) \cite{UR3_ref} is sourced from Universal Robots, which has 6 DOF and has a payload capacity of 3 kg.  There is a two finger gripper from Robotiq (2F-85 gripper) to hold objects. Apart from this, the ROS2 API software is installed in the PC and transmits control commands to the real robot. The hardware in loop 1 and the hardware in loop 2 are networked using IP over Ethernet protocol.

\subsection{Digital twin construction}
The remote area, including the real robot, is modeled as a digital twin with which the operator interacts. The simulated environment is expected to be realistic to provide an immersive experience for operators in the real-time teleoperation setting. We chose NVIDIA's Isaac Sim, a powerful simulation platform designed for robotics research and development. We leverage the visual realism and physics engine of Isaac Sim to give the human operator an experience of being physically present on the remote side.




To create the digital twin, we imported the Unified Robot Description Format (URDF) file of the UR3 robotic arm and Robotiq 2F-85 gripper into the current simulation scene. 
URDF is an XML file format that describes a robot’s elements, such as links and joints.
The data contained in URDF files are used to model multi-body systems such as robotic arms and provide information about their physical description, including joints, motors, and mass.
The URDF files are imported inside the Isaac Sim and then converted into a single Universal Scene Description (USD) file, where they are both connected to each other by a fixed joint. This USD is used for loading the scene in the simulation.
USD is an open and extensible ecosystem for describing, composing, simulating, and collaborating within 3D worlds.
Alongside this, we utilized a robot description file generated using the Lula Robot Description Editor. This file defines the robot's actuated joints for motion planning and represents its collision geometry using spheres for collision detection.
 We customized the background color and ambient lighting to closely resemble the physical environment, creating a more immersive and lifelike representation. 

To calibrate and synchronize all the systems in space, the real robot, the simulated robots, and the haptic device are pose corrected. When the haptic device stylus is in the home position, we enforce the pose, and thus the orientation difference between the simulated and real robot is zero. This implementation will ensure that the movement of the stylus, simulated robots, and the real robot are space synchronized.


\subsection{Need for a second digital twin}

Effective teleoperation relies on immersing the operator in the remote environment. Traditional video-based teleoperation provides a visual representation; however, additional depth information is crucial for effective control. For a truly immersive environment, multiple sensors (visual and depth) must be deployed across the entire remote environment from all angles. However, as the size of the workspace increases, the number of required sensors also increases, and streaming data from these sensors in real-time introduces significant bandwidth and network challenges. To address these challenges, we propose a novel approach that minimizes data transfer. Rather than streaming all sensor data, we introduce a second digital twin, referred to as DT Robot 2, located on the remote side. It mirrors DT Robo1 1 and will facilitate the transfer of operator side essential control data such as the robot's end-effector position and status (x, y, z, o, c). Now, since the digital twin is already at the remote side, we compare the DT Robot 2 and the real environment to find the discrepancies through the methodology discussed in Section~\ref{section E}, and transfer this condensed important data back to the operator. Thus, this significantly reduces network traffic and enables more efficient teleoperation while addressing the challenges associated with large-scale environments.

\begin{figure}[t]
    \centering
    \includegraphics[width=0.49\textwidth]{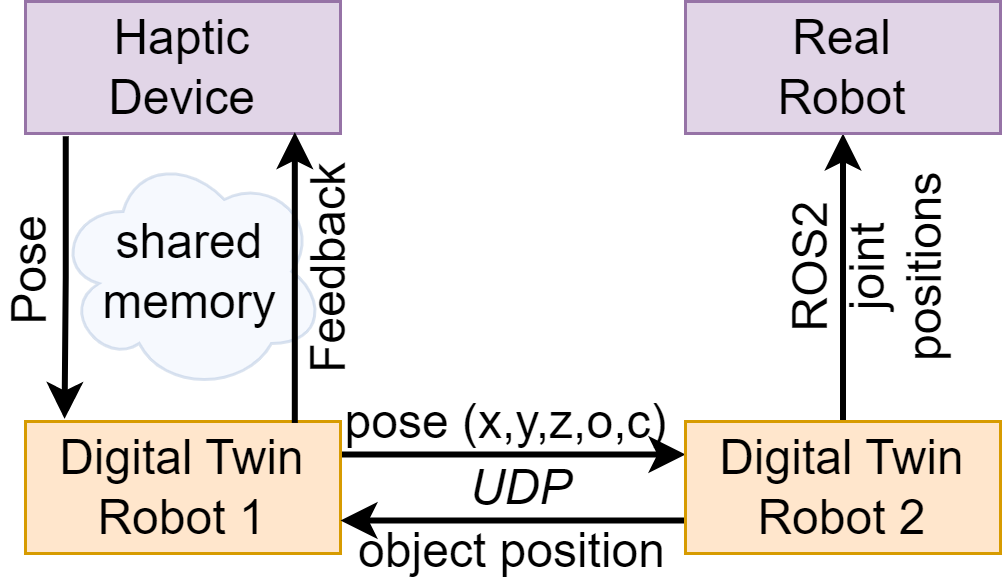}
    \caption{Data Flow}
    \label{loops}
\end{figure}

\subsection{Loop 1: Human Side and Digital Twin}

Loop 1 shown in Fig. \ref{loops1} comprises two main components: the user controlling the robot within the simulated environment and the digital twin providing haptic feedback to the user. The operator interacts with DT Robot 1 using the haptic device. The digital twin exposes a green sphere (as seen in Fig. \ref{fig:image3}) for the user to move to a desired position 
using the haptic device.  The data from the haptic device is continuously written to a shared memory, which the digital twin continuously reads, as shown in Fig. \ref{loops}, and computes the respective joint positions of the simulated robot for its end effector to reach the desired position by applying an inverse kinematic algorithm. In our work, we used the RMPFlow algorithm \cite{rmp_flow} for inverse kinematic calculations. Riemannian Motion Policy (RMP) is a framework for smooth, geometry-aware motion generation in robots, combining acceleration-based policies with Riemannian metrics. RMPflow extends this by organizing multiple local policies across various task spaces into a single, consistent global policy using principles from Riemannian geometry. It enables intelligent behaviors like goal-reaching and reactive collision avoidance by prioritizing safe, efficient paths around obstacles.


In the second component, feedback is essential to prevent the user from pushing the robot too hard against any surface. This feedback from the digital twin is obtained through contact sensors placed at the tip of the tool held by the end effector. Contact sensors continuously publish the forces they detect to shared memory. The Geomagic Touch device constantly reads this memory and applies force feedback to the user proportionally to the readings from the contact sensors.
  

\subsection{Loop 2: DT Robot 1 and remote side} \label{section E}
The DT Robot 2 in loop 2 has to accomplish two main functions: (1) ensuring that the real physical robot follows the motion of its digital counterpart in DT Robot 1 and (2)  updating the environment of the DT Robot 1 whenever there are changes from the real environment caused by abrupt changes that occurred due to external disturbances, as well as imperfections in the modeling of the twin.

DT Robot 1 transmits the pose 
through the UDP protocol as shown in Fig. \ref{loops}. DT Robot 2 receives this data and computes the trajectory using RMPflow. ROS2 is used to communicate the calculated joint positions and gripper state to the real robot. The real robot uses a scaled joint trajectory controller to execute joint-space trajectories on a group of joints by interpolating in time between the points so that their distance can be arbitrary. In our ROS2 implementation, we use the subscriber `fire and forget' mechanism to send the joint trajectory to the controller. While this does not explicitly offer positional feedback, our force feedback is sufficient. Moreover, since joint positions are published at 100 Hz, an isolated data loss will not hurt the trajectory, as the robot moves smoothly to the new position by concealing the loss. 

\begin{figure}[h!]
    \centering
    \begin{subfigure}[b]{0.3\textwidth}
        \centering
        \includegraphics[width=\textwidth]{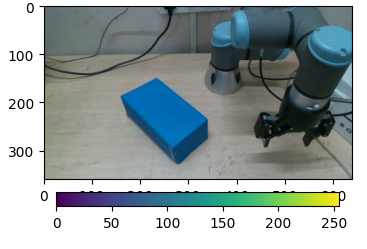 }
        \caption{RGB from real camera}
        \label{fig:image1}
    \end{subfigure}
    \begin{subfigure}[b]{0.3\textwidth}
        \centering
        \includegraphics[width=\textwidth]{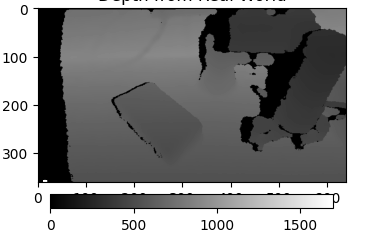 }
        \caption{Depth from real camera}
        \label{fig:image2}
    \end{subfigure}
    \begin{subfigure}[b]{0.3\textwidth}
        \centering
        \includegraphics[width=\textwidth]{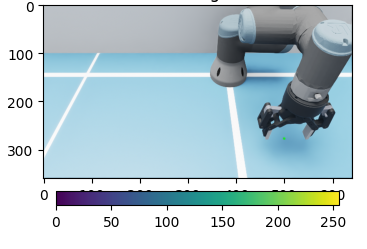 }
        \caption{RGB from simulated camera}
        \label{fig:image3}
    \end{subfigure}
    \begin{subfigure}[b]{0.3\textwidth}
        \centering
        \includegraphics[width=\textwidth]{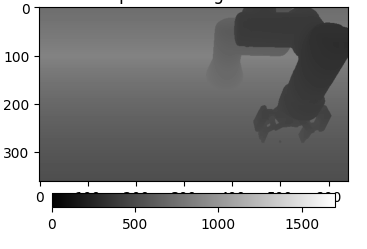 }
        \caption{Depth from simulated camera}
        \label{fig:image4}
    \end{subfigure}
    \begin{subfigure}[b]{0.3\textwidth}
        \centering
        \includegraphics[width=\textwidth]{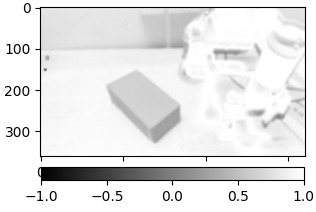}
        \caption{Output after SSIM}
        \label{fig:image5}
    \end{subfigure}
    \caption{Images from both cameras, and image after applying SSIM}
    \label{fig:six_images}
\end{figure}

The second function is to update DT Robot 2 with the changes from the real world. For capturing the real world scene around the real robot, we utilise the Intel RealSense depth camera D415 \cite{RealSense_ref}. Let us call this a real camera feed. We create a digital replica of this camera with the same specifications by setting the focal length, aperture (horizontal, vertical), and other intrinsic parameters. We then place the digital camera in the same position and orientation as its real-world counterpart. We call this 
 the synthetic camera feed.  Fig. \ref{fig:image1}, \ref{fig:image2}, \ref{fig:image3}, \ref{fig:image4} present sample RGB and depth images from both real and virtual cameras.

\begin{figure}[h!]
    \centering
    \begin{subfigure}[b]{0.35\textwidth}
        \centering
        \includegraphics[width=\textwidth]{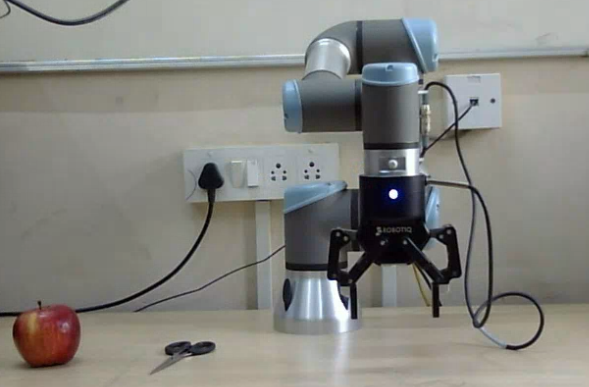}
        \caption{Object detection from real camera}
        \label{fig:yolo_in_real}
    \end{subfigure}
    \hfill
    \begin{subfigure}[b]{0.35\textwidth}
        \centering
        \includegraphics[width=\textwidth]{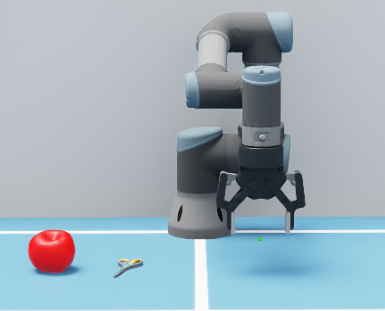 }
        \caption{Objects rendered in simulation}
        \label{fig:yolo_in_dt}
    \end{subfigure}
    \hfill
    \begin{subfigure}[b]{0.35\textwidth}
        \centering
        \includegraphics[width=\textwidth]{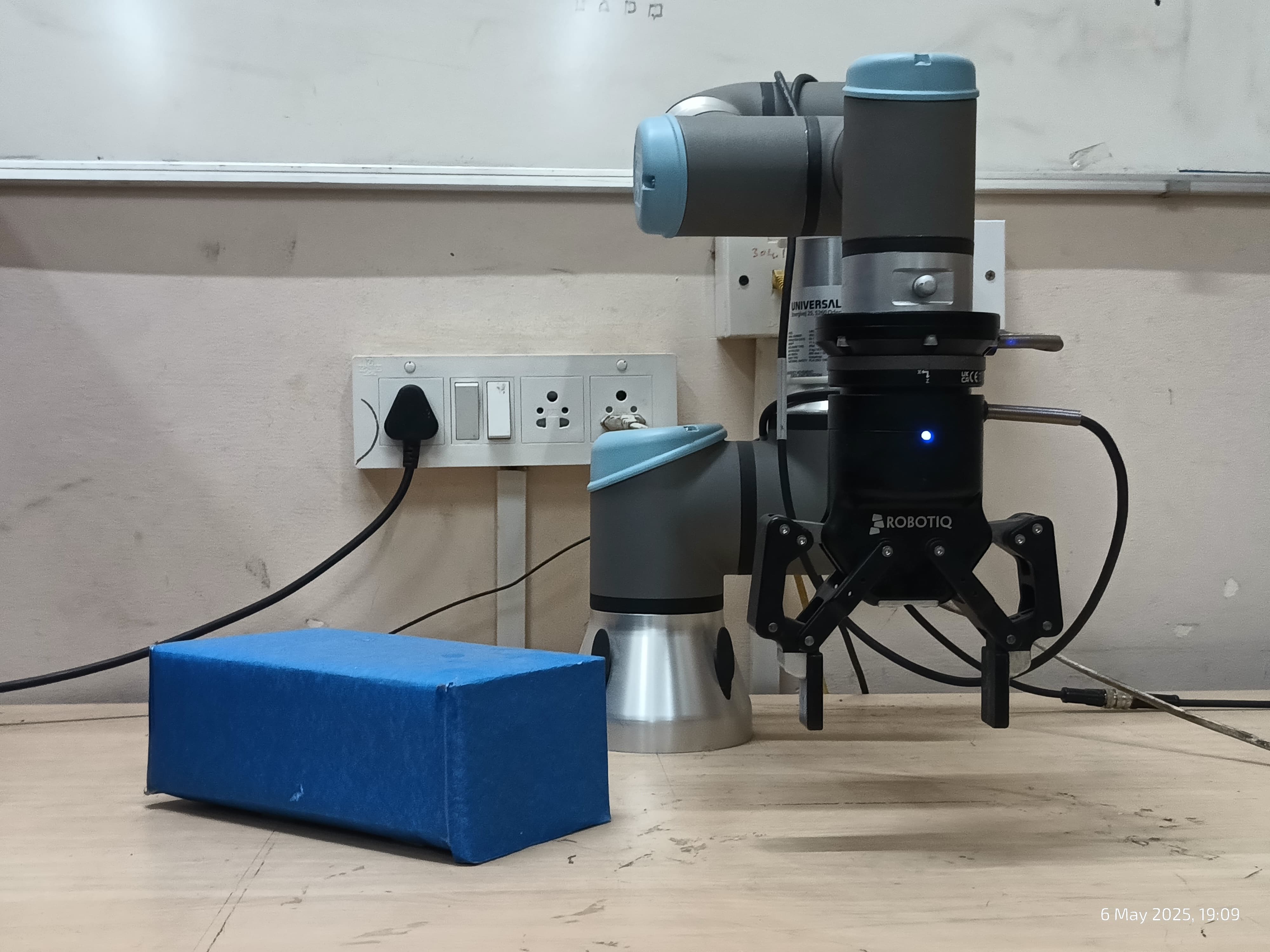}
        \caption{Foreign objects}
        \label{fig:foreign_objects}
    \end{subfigure}
    \hfill
    \begin{subfigure}[b]{0.35\textwidth}
        \centering
        \includegraphics[width=\textwidth]{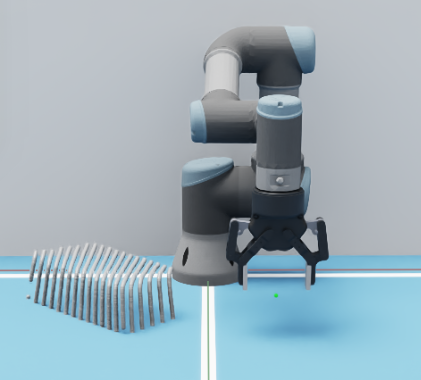}
        \caption{Foreign objects as point cloud}
        \label{fig:point_cloud}
    \end{subfigure}
    \caption{Object rendering inside the simulation}
    \label{fig:object_render}
\end{figure}

\textbf{Working with Known Objects }: In the real-world scene, surgery is performed systematically using surgical tools such as scalpels, scissors, forceps, clamps, retractors, and suction devices. These tools are used to accomplish multiple functions, such as cleaning, cutting, grasping, suturing, etc. In our architectural framework, we utilize deep learning-based AI software to detect and identify these well-known objects (see Fig. \ref{fig:yolo_in_real}). We used YOLOv8 \cite{yolov8_ultralytics} to identify the object as well as its \text{(x, y, z)} coordinates in the scene, and using this information, we render the identified object inside the simulated environment (see Fig. \ref{fig:yolo_in_dt}).

In our experimental setup, since we use a single 3D camera, there can be several blind spots where the tools can get occluded. We enlist the following scenarios:
1. Robot occluding the object: if the robot comes between the camera and the object, then the YOLOv8 object detection algorithm will fail.
2. Object at the boundary: If the object is placed at or near the boundary of the real-world scene, the detection accuracy of YOLOv8 decreases.
3. Object close to camera sensor: if the minimum distance to the camera is violated, then the YOLOv8 object detection algorithm will fail.
While our current setup utilises a single 3D camera, our proposed architecture is sufficiently modular to install multiple 3D cameras to overcome blind spots.

\RestyleAlgo{ruled}

\SetKwComment{Comment}{/* }{ */}
\SetCommentSty{mycommfont}
\begin{algorithm}[t!]
\label{alg:dis}
\caption{Discrepancy Detection and Integration}\label{alg:DiscrepancyDetection}
\KwData{RGB and depth image from real-world ($RGB_{R}$, $D_{R}$), Synthetic RGB and synthetic depth image from digital twin ($RGB_{S}$, $D_{S}$)}
\KwResult{Discrepancy Point Cloud ($PC$)}

    \Comment{Compute Structural Similarity Index Measure (SSIM)}
    $SSIM_{RGB} \gets $ Compute\_SSIM($RGB_{R}$, $RGB_{S}$)\;
    $SSIM_{D} \gets $ Compute\_SSIM($D_{R}$, $D_{S}$)\;

    \Comment{Combine RGB and Depth Signals}
    $FusedImage \gets \min(SSIM_{RGB}, SSIM_{D})$ \;

    \Comment{Apply Morphological Operations}
    $ProcessedImage \gets $ Erode($FusedImage$)\;
    $ProcessedImage \gets $ Dilate($ProcessedImage$)\;

    \Comment{Extract Discrepancy Data as Point Cloud}
    $PC \gets $ Extract\_PointCloud($ProcessedImage$)\;

    \Comment{Integrate into DT Robot 2 and Transmit to DT Robot 1}
    Integrate\_PointCloud($PC$, Digital Twin 2)\;
    Transmit\_Over\_UDP($PC$, Digital Twin 1)\;
    
\end{algorithm}

\textbf{Working with foreign objects} :
Since performing surgery on humans can throw up unexpected outcomes, the remote side can introduce equipment and utilise new tools that are foreign to trained YOLOv8(as shown in Fig. \ref{fig:foreign_objects}). Also, when YOLOv8 detects and identifies objects with lower probabilities (less than 0.9), we require a discrepancy detection algorithm to detect objects present in the real-world scene.
Algorithm \ref{alg:DiscrepancyDetection} outlines the discrepancy detection method. We employ the Structural Similarity Index Measure (SSIM) on both RGB and depth data to identify inconsistencies (see Fig. \ref{fig:image5}). We use the simulated RealSense D415 camera inside Isaac Sim as a reference synthetic feed. Foreign object detection is accomplished when the real image from the real camera and the synthetic image from the simulated camera are superimposed onto each other to find the pixel to pixel difference. After we obtain the difference, the coordinates of these differences are gathered from the RealSense depth data to create a point cloud and render it inside the Isaac Sim. While depth data alone would ideally suffice for discrepancy estimation, Fig. \ref{fig:image2} illustrates the presence of noise in real-world depth images, necessitating the fusion of RGB and depth signals. We integrate these signals by computing a pixel-wise minimum. To further mitigate small artifacts introduced by noise, morphological operations such as erosion and dilation are applied. The extracted discrepancy data is then integrated into DT Robot 2 as a point cloud, as shown in Fig. \ref{fig:point_cloud}. This information is subsequently transmitted to DT Robot 1 over UDP (see Fig. \ref{loops}), allowing the user to visualise in real time.







\section{Features of RoboTwin}
\subsection{Using a single RBG-D camera}
Typically, to perceive the size and position of an object in 3D space, we require 3 cameras, catering to each dimension. While more cameras can improve accuracy, our approach is minimalistic and can also act as a fallback option when one or more of the cameras fails.
In this work, we use a single RealSense \cite{RealSense_ref} camera with RGB and depth sensing mounted at an angle to sense and estimate the pose of an object.

\subsection{Safety}

Conventional robot teleoperation typically relies on visual feedback from cameras and basic force sensing. In a study by Pal et al. \cite{pal2024testbed}, video feedback latency was measured as 44ms over a distance of $400~km$. To address operational safety concerns, they implemented local haptic feedback using basic boundary conditions. However, this approach limits the operator's perception of complex interactions with the remote environment.
Our approach leverages the fully modeled remote environment within Isaac Sim to provide low-latency visual and haptic feedback. By replicating surfaces and objects as a digital twin, we enable tangible interaction through force and visual cues. This makes the teleoperation intuitive to the operator. Furthermore, Isaac Sim can prevent unsafe actions by either not transmitting the robot pose or modifying it before execution. This safeguards the teleoperator and the real environment. Essentially, the DT Robot 2 acts as a protective layer before ROS2 commands reach the real robot.

\subsection{Motion Scaling}
Tasks involving tiny movements, such as the case of microsurgery, can be severely affected by physiological tremors of the hands. This is especially true when the surgeon operates sensitive equipment such as the Geomagic haptic device. The authors in EdgeP4 \cite{gnani2024edgep4} showed that in-network filtering of tremors is done using a threshold based filter. However, this also suppresses intentional micro movements, particularly required in brain surgeries. To overcome this limitation, we apply a scale-down between the movement of the haptic device and the DT Robot. To clearly view the micromovements, we zoom into the target area in the digital twin, as this improves the accuracy and precision of the intended motion. We also apply scale-up as it might be a requirement to perform macromovements. Such examples include large sized sutures to arrest blood loss. We explored the possibility of working with a digital twin to accomplish micro and macro movements.

\section{Experimental Design And Results}
Our study focuses on the NASA Task Load Index (NASA-TLX) \cite{nasatlx_ref} metric when operators work on the digitally rendered images and video. The goal is to bridge the wide gap that currently exists between the novice, intermediate, and expert operators.
NASA-TLX is a widely used subjective workload assessment tool designed to evaluate the cognitive and physical demands of a task. It measures workload across six dimensions: mental demand, physical demand, temporal demand, performance (shortcomings), effort, and frustration. In our user study, the participants rated each factor on a 1-7 scale. In each parameter, the lower the score, the better is user experience. We consider the spiral drawing task~\cite{SENKIV2019190} as it involves intricate handling of the haptic device.
To evaluate the performance of the RoboTwin framework, we conducted a user study with 17 participants, aged 20 to 35 years, of various backgrounds. Each participant performed a remote spiral drawing task using the haptic device, which is used to control the robotic arm equipped with a marker pen. A spiral printed on paper was placed in front of the robot, and the task required tracing the spiral while staying within the printed path. The testbed includes an RGB depth camera and a software tool to introduce motion scaling and network latency. A few known and foreign objects are introduced. The experimental study for the operator included (a) camera-based visual and haptic feedback from the real world, as well as (b) the digital twin. The round-trip-time (RTT) network latency we considered was 1 ms and 100 ms throughout the task.

\subsection{Workload Assesment}
\begin{figure}[]
    \centering
    \includegraphics[width=0.48\textwidth]{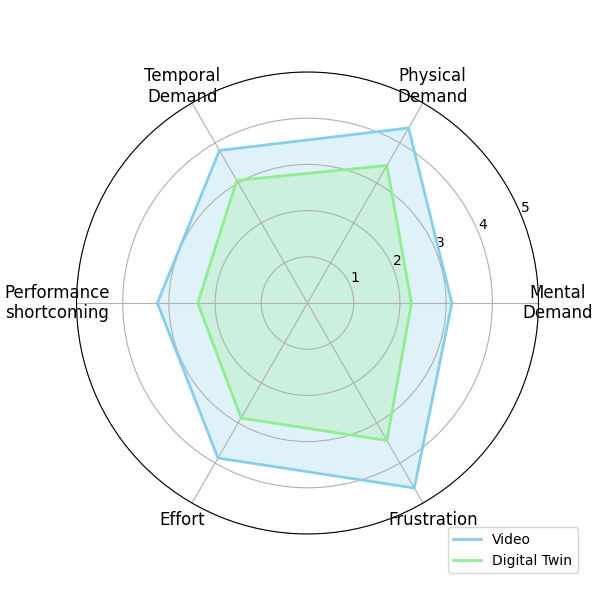}
    \caption{Average of the NASA TLX scores by participants (lower is better)}
    \label{fig:tlx_avgSpider}
\end{figure}
Fig. \ref{fig:tlx_avgSpider} summarizes the cognitive and physical demand experienced by the participants for 1 ms as well as 100 ms RTT. This radar chart visualizes the NASA-TLX ratings for the spiral-tracing task under two different visual feedback conditions: (a) video (shown in blue colour) and (b) digital twin (shown in green colour). 
The six workload dimensions are plotted on the axes. The results indicate that the video-based interface imposes a significantly higher workload across all dimensions, particularly in physical demand, frustration, and effort, suggesting that users found it more challenging to control the robotic arm accurately. In contrast, the digital twin condition consistently shows lower workload ratings, indicating that it provides a more intuitive and manageable interface for teleoperation. These findings align with the spiral-tracing results presented in the Section \ref{subsec:spiral_compare}, where digital twin feedback facilitated more precise movements, even with 100 ms network latency.

\begin{figure}[]
    \centering
    \includegraphics[width=0.48\textwidth]{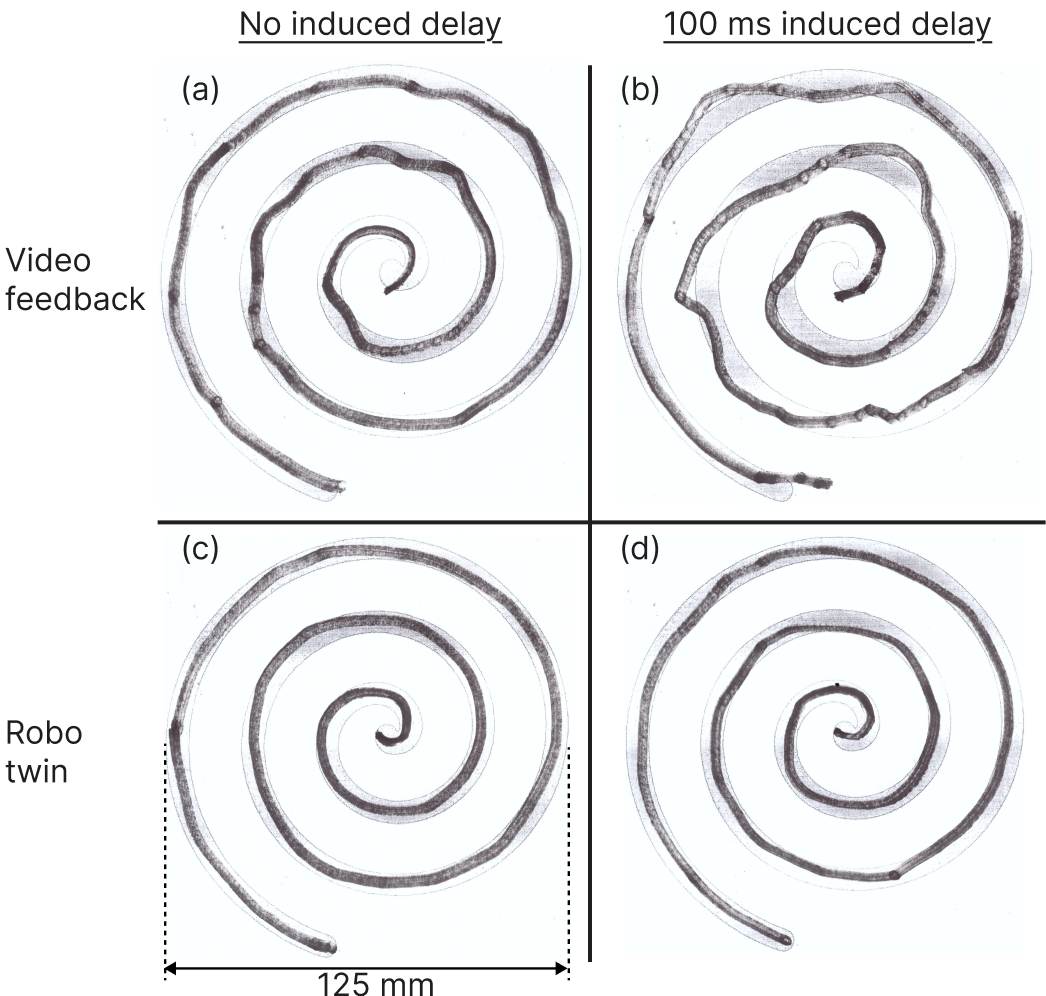}
    \caption{Spirals drawn under different scenarios}
    \label{fig:spiral_compare}
\end{figure}

\subsection{Task completion quality} \label{subsec:spiral_compare}
We now compare the task-completion quality and present the results of the spiral-tracing task under different conditions as shown in  Fig. \ref{fig:spiral_compare}. It comprises of spirals drawn using the two scenarios that compare the effect of real camera video-feedback with RoboTwin feedback under the two RTT network latencies. The top row, (a) and (b), corresponds to tasks performed using video feedback, while the bottom row, (c) and (d), represents results obtained with the RoboTwin visualization. The left column shows the performance under no induced network delay, whereas the right column illustrates the impact of an induced 100 ms RTT delay. The results indicate that using video feedback for long distance teleoperation (Fig. \ref{fig:spiral_compare}b), leads to numerous straying off the path and overshoots, indicating loss of trajectory control. In contrast, the RoboTwin visualization (Fig. \ref{fig:spiral_compare}d) enables more stable and accurate tracing, even under extreme delay conditions. This suggests that the RoboTwin provides a more reliable interface for teleoperation by mitigating the adverse effects of network latency.

While the task completion time for the RoboTwin-based visualization tasks (Fig. \ref{fig:spiral_compare}c, \ref{fig:spiral_compare}d) is the same, another significant benefit is that the video feedback data from the real robot to RoboTwin is practically zero, although there is haptic feedback. However, for the video-feedback based tasks (Fig. \ref{fig:spiral_compare}a, \ref{fig:spiral_compare}b) the task completion time of (Fig. \ref{fig:spiral_compare}b) is longer. Also, since the task in Fig. \ref{fig:spiral_compare}b represents 100 ms RTT latency, a 100 ms increase in video-feedback latency is observed. 

\subsection{Motion Scaling}
The robotic surgery experts suggested that for intricate surgeries, the gain of the system, which involves motion scaling, should be implemented. For our experiments, we chose a 30\% decrease and increase from the normal speed of an expert operator. At normal speed, a 1 mm movement of the haptic device corresponds to a 1mm movement of the robotic arm (digital twins as well as real robots). For micromovements, a 1.3mm movement of the haptic device corresponds to 1 mm of the robot, and similarly for macromovements, 
a 0.7mm movement of the haptic device corresponds to 1mm of the robot.

\begin{table}[h!]
\centering
\renewcommand\cellalign{lc}
\renewcommand\cellgape{\Gape[5pt]}
\begin{tabular}{|c|c|c|c|}
\hline
\makecell{Movement} & \makecell{Completion\\Time (sec)} & \makecell{Bytes\\Transmitted (KB)} & \makecell{Image} \\
\hline
Macro & 31.69 & 72.9 & \includegraphics[width=0.9cm]{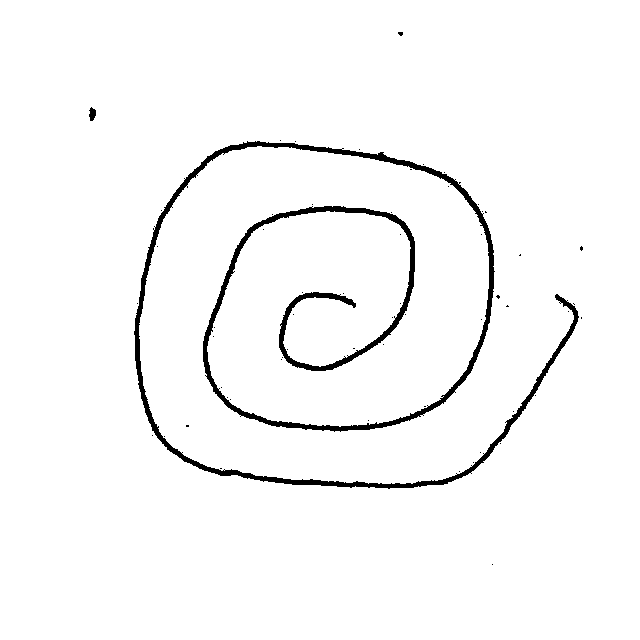} \\
\hline
Normal & 48.08 & 106.4 & \includegraphics[width=0.9cm]{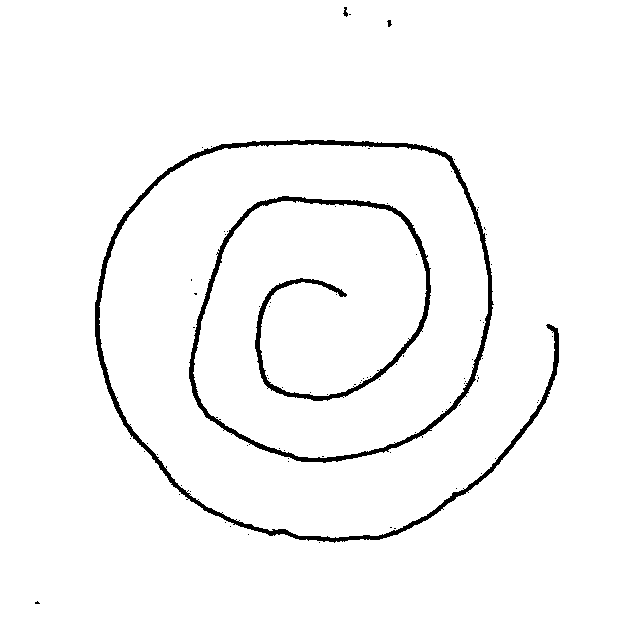} \\
\hline
Micro & 65.51 & 145.1 & \includegraphics[width=0.9cm]{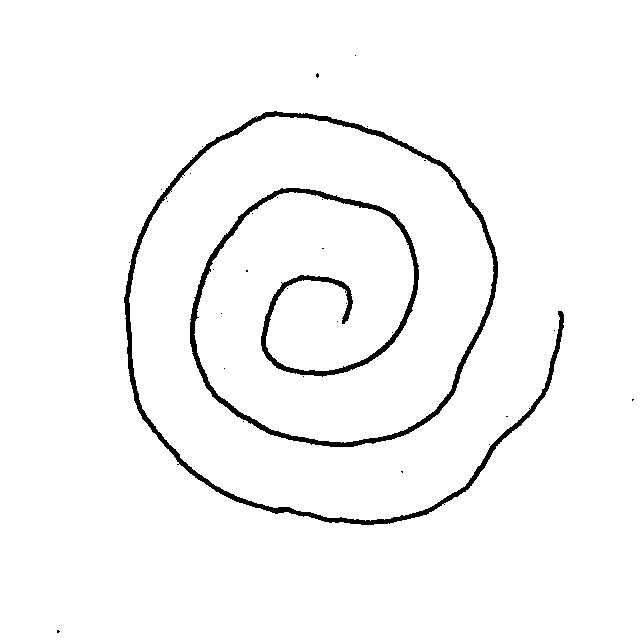} \\
\hline
\end{tabular}
\caption{Motion scaling}
\label{tab:motion_scaling}
\end{table}

Table \ref{tab:motion_scaling} shows the results of motion scaling when an expert operator draws the spiral. Since the sampling frequency of the haptic device is 1 kHz, the amount of data transmitted will be based on speed of completion. For micromovements, the expert operator completes the task in roughly 35\% more than the normal time, and for macromovements, the time reduction is 35\%. Similarly, data transferred in case of micromovements is 33\% more than normal, and in case of macromovements, it is 32\% less than normal.
As for the task-completion quality, the image corresponding to the micro movements is superior at the cost of 35\% more time than the normal. 

\begin{figure}[h!]
    \centering
    \begin{subfigure}[t]{0.49\textwidth}
        \centering
        \includegraphics[width=\textwidth]{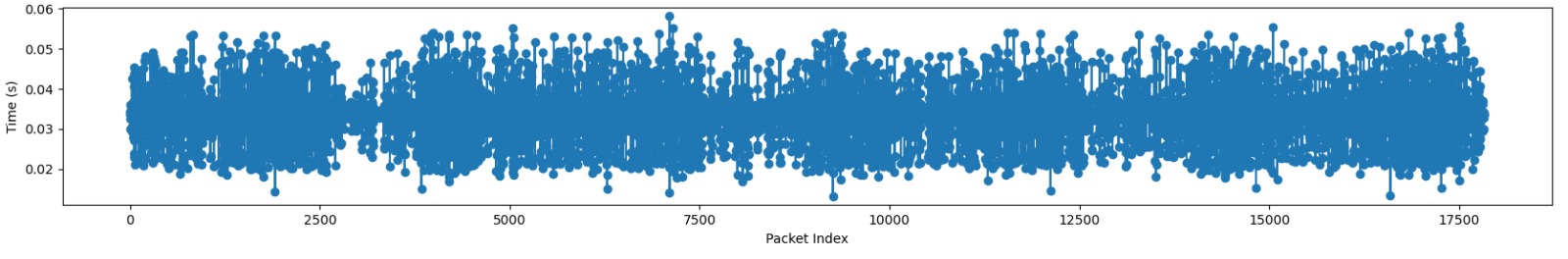}
        \caption{video-based rendering}
        \label{fig:image7}
    \end{subfigure}
    \hfill
    \begin{subfigure}[t]{0.49\textwidth}
        \centering
        \includegraphics[width=\textwidth]{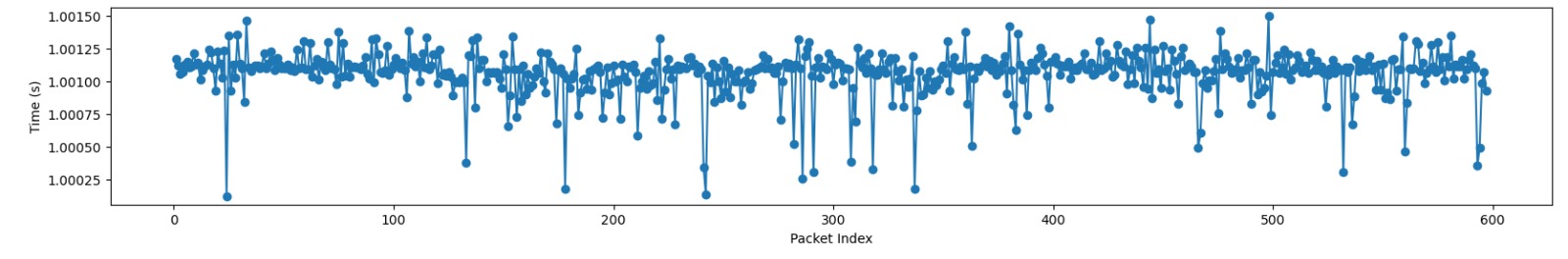}
        \caption{coordinate-based object rendering}
        \label{fig:image8}
    \end{subfigure}
    \caption{Bandwidth comparison for rendering objects for two different approaches}
    \label{fig:bandwidth_comp}
\end{figure}

\subsection{Bandwidth Requirement}
Let us now turn our attention to the bandwidth requirement for our RoboTwin architecture. In Fig. \ref{fig:image7}, the x-axis shows that 17,500 packets of size 1,500 bytes are required to be transferred to the operator side for rendering known objects using the conventional approach. The y-axis shows the inter-arrival time of the packets on the operator side. In essence, this captures the bandwidth requirement. Whereas, in Fig. \ref{fig:image8}, x-axis shows that only 600 packets of size 46 bytes are transferred, corresponding to the x, y, and z coordinates of the object as well as fulfilling ethernet's minimum frame size. The y-axis shows that the bandwidth requirement is 25 times lower than the conventional approach for full rendering of the video.



\section{Conclusion}In this work, we have proposed, implemented, and tested a novel architecture for remote robotic surgeries. This dual loop architecture comprises operators interacting locally with the digital twin, with bidirectional traffic flow between the operator side and the remote side while overcoming the cyber-sickness issues associated with large round-trip latencies. Experimental results show that the architecture benefits from good task completion quality as well as user experience. The architecture is very well suited for bandwidth constrained internet connections without compromising the quality of surgery. The testbed architecture has several features, such as a geofenced operating area, real time object detection (known and foreign objects) for safety purposes, as well as motion scaling for micro and macro surgical movements.






\bibliographystyle{IEEEtran}
\bibliography{IEEEabrv,references} 

\end{document}